\documentclass[sigconf]{acmart}
\usepackage{amsmath,amsfonts,bm}

\newcommand{\buyers}{{{\ensuremath{B}}}}
\newcommand{\sellers}{{{\ensuremath{S}}}}
\newcommand{\items}{{{\ensuremath{U}}}}
\newcommand{\recommendations}{{{\ensuremath{\mathcal{R}}}}}

\DeclareMathOperator*{\argmax}{arg\,max}

\newcommand{\etal}{\emph{et al.}}

\newcommand{\squishlist}{\begin{list}{$\bullet$}
  { \setlength{\itemsep}{0pt}
     \setlength{\parsep}{3pt}
     \setlength{\topsep}{3pt}
     \setlength{\partopsep}{0pt}
     \setlength{\leftmargin}{1.5em}
     \setlength{\labelwidth}{1em}
     \setlength{\labelsep}{0.5em} } }
\newcommand{\squishend}{
  \end{list}  }

\newtheorem{problem}{Problem}
\usepackage{mathtools}
\usepackage{amsmath}
\usepackage{amssymb}
\usepackage{amsthm}
\DeclarePairedDelimiter{\ceil}{\lceil}{\rceil}
\usepackage{subfig}
\usepackage[noend, linesnumbered, ruled]{algorithm2e} % For algorithms

\SetKwIF{If}{ElseIf}{Else}{if}{}{else if}{else}{end if}%
\SetKwFor{While}{while}{}{end while}%
\SetKwFor{For}{for}{}{end for}
\SetKwRepeat{Do}{do}{while}
\AtBeginDocument{%
  \providecommand\BibTeX{{%
    \normalfont B\kern-0.5em{\scshape i\kern-0.25em b}\kern-0.8em\TeX}}}

\setcopyright{none}
\copyrightyear{2020}
\acmYear{2020}

\acmConference[RecSys '20]{FAccTRec Workshop: Responsible Recommendation}{September 26, 2020}{Worldwide}
\begin{document}

\title{A General Framework for Fairness in Multistakeholder Recommendations}

\author{Harshal A. Chaudhari}
\authornote{Work performed during an internship at Zillow Group.}
\affiliation{%
  \institution{Boston University}
  \city{Boston, MA}
}
\email{harshal@cs.bu.edu}

\author{Sangdi Lin}
\affiliation{%
  \institution{Zillow Group}
  \city{Seattle, WA}
}
\email{sangdil@zillowgroup.com}

\author{Ondrej Linda}
\affiliation{%
  \institution{Zillow Group}
  \city{Seattle, WA}
}
\email{ondrejl@zillowgroup.com}

\renewcommand{\shortauthors}{Chaudhari, Lin and Linda.}

%% Paper content
\begin{abstract}
Contemporary recommender systems act as intermediaries on multi-sided           
    platforms serving
    high utility recommendations from sellers to buyers.
Such systems attempt to balance the objectives of multiple stakeholders including
    sellers, buyers, and the platform itself.
The difficulty in providing recommendations that maximize the utility for a buyer,  
    while simultaneously representing all the sellers on the platform 
    has lead to many interesting research problems.
Traditionally, they have been formulated as integer linear programs which
    compute recommendations for all the buyers together in an \emph{offline} fashion, by 
    incorporating coverage constraints so that the individual sellers
    are proportionally represented across all the recommended items.
Such approaches can lead to unforeseen biases wherein certain buyers consistently 
    receive
    low utility recommendations in order to meet the global seller coverage constraints.
To remedy this situation, we propose a general formulation that incorporates seller 
    coverage objectives alongside individual buyer objectives in a real-time
    personalized recommender system.
In addition, we leverage highly scalable submodular optimization algorithms 
    to provide recommendations to each buyer with provable theoretical quality bounds.
Furthermore, we empirically evaluate the efficacy of our approach using data from 
    an online real-estate marketplace.
\end{abstract}
\maketitle

\section{Introduction}

The rise of e-commerce platforms in the past decade have made recommender systems 
    ubiquitous over the world wide web.
Recommender systems typically assist the buyers on a web marketplace
    by recommending them items that are closely aligned to their preferences, 
    thereby significantly
    reducing the time required for search.
They have been successfully used in several different domains viz., e-commerce 
    platforms such 
    as Amazon, eBay, etc., media streaming platforms like Netflix, Spotify, etc., social
    networks like Facebook, Twitter, etc., as well as the hospitality services like Yelp, Airbnb, etc.
    
Traditionally, such systems have always aimed at maximizing the utility of 
    recommendations by tailoring them towards the preferences of an individual 
    target buyer.
Such recommendations are referred to as \emph{personalized recommendations}.
Increasingly, the advent of multi-sided marketplaces such as Airbnb, UberEats, etc. 
    have shone spotlight on the issue of welfare of other stakeholders, who are also
    affected by these buyer-oriented recommender systems.
Multi-sided marketplaces, which primarily rely on \emph{network effects} for growth 
    are therefore increasingly motivated to include their objectives in addition to
    buyers.
Providing meaningful exposure to new sellers or niche brands that are attractive to 
    small market segments, supporting small businesses as they compete
    with the conglomerates for buyer attention, etc. are a few 
    objectives important to the other stakeholders of the platforms.
Without explicitly accounting for such goals, the recommender 
    systems can cause undesirable biases, filter bubbles, and contribute to the
    `rich become richer' phenomenon on their platforms.
    
In this work, we propose a scalable multi-stakeholder
    recommender system capable of optimizing for multiple criteria
    across different stakeholders.
Specifically, we consider individual sellers on the platform as different
    stakeholders, who would like their items to be
    proportionally represented in the recommendations.
This problem has traditionally been formulated as an integer linear
    program~\citep{Surer2018-bh, Malthouse-undated-vs}.
However, the heuristic algorithms used to solve such integer programs cannot provide
    guarantees on the quality of solution when compared to the optimal solution.
In contrast, we formulate the task as a multi-objective optimization problem 
    consisting of submodular
    stakeholder coverage objective augmented with linear (modular) 
    auxiliary objective. 
This task is solvable in a computationally
    efficient manner while also providing provable guarantees on the quality of the solution.   
The main contributions of our work are:
\squishlist
    \item Formulation of fair multi-stakeholder recommendations as a     
        submodular maximization problem, capable of incorporating multiple auxiliary objectives simultaneously, while providing strong approximation guarantees
        on the quality of solution.
    \item Empirical evaluation using data from Zillow, a real estate marketplace.
\squishend
\vspace{-4pt}
\section{Background}

In addition to extensive literature on personalized recommendations, 
    multi-receiver/multi-provider recommendations, etc., in recent years, there is a growing interest in analyzing
    recommendation systems under the lens of \emph{fairness}.
In this section, we position our work in the context of broad
    recommender systems and related optimization techniques.
    
\subsection{Multistakeholder Systems}
Multistakeholder recommender systems are a broad category of recommender systems 
    that involve more than one stakeholders.
In their simplest form, the reciprocal recommendation systems including 
    `person-to-person' 
    links on social networks~\cite{Guy2015-aa},
    online dating~\citep{Xia2015-cj} % Pizzato2010-kc} 
    and job search platforms~\cite{Mine2013-af} are all examples of systems with two stakeholders.
We focus on large scale multi-sided platforms such as Amazon, Alibaba, Airbnb, etc.,
    connecting sellers to buyers.
%     where multiple independent sellers of varying sizes are competing for attention of buyers.
% At the same time, Amazon's \emph{in house brands} compete with some of the sellers 
%     while also providing higher profit margins, and therefore need to be prioritized
%     in order to serve the best interests of the platform's investors.
Ideally, the percentage of items belonging to each seller in the items recommended
    to a buyer should be proportional to the number of items belonging to the particular stakeholder that are relevant to the buyer.
    
Solutions to the above challenges using  multi-objective optimization are explored
    in the works such as \citep{Burke2016-vw, Abdollahpouri2017-kx}.
    %Ribeiro2012-dz}. 
The importance of price and profit awareness in the recommender systems is studied
    by \cite{Jannach2017-pf} and \cite{Pei2019-hb}. 
Works like \citep{Yao2017-sj, Modani2017-pa, liu2018personalizing}
% Serbos2017-gw} 
contribute to the very
    important domain of analyzing the fairness of recommender systems.
Ekstrand and Kluver~\cite{ekstrand2018exploring} explore the gender-discriminatory effects of 
    collaborative filtering
    in book ratings and recommendations.
Works like ~\cite{sonboli2019localized} and ~\cite{kamishima2018recommendation} explore the impact
    of sensitive variables on the fairness of recommendations systems.
Perhaps closest to our objective, the recent work of S{\"u}rer {\etal}
    \cite{Surer2018-bh}
    imposes fairness constraints in terms of minimum coverage of different sellers
    across recommendations to all the buyers using sub-gradient based methods to reformulate the
    coverage optimizing integer program.
In contrast, our formulation does not impose strict coverage constraints. 
Intuitively, in situations where adhering to strict coverage constraints results
    in dramatic sacrifice of utility of recommendations, our formulation 
    automatically relaxes such constraints while still satisfying strong 
    approximation bounds in regards to quality of solution.
Moreover, our formulation determines recommendations per buyer and can be
    incorporated into any personalized recommendation system in form a
    post-processing step.
    
\subsection{Optimization Techniques}
This section reviews some of the recent advances in the field of submodulaar
    optimization that our method heavily relies on. Filmus and
    Ward~\cite{Filmus2012-py}  provided one of the earliest methods to maximize a 
    monotone submodular function under matroid constraints.
Leveraging multi-linear relaxations of submodular functions,
    Feldman~\cite{Feldman2018-vi} proposed
    a continuous greedy algorithm that approximately maximizes an objective 
    comprised of a submodular function and an arbitrary linear function.
Recently, scalable greedy algorithms with similar approximation bounds, to 
    maximize difference between a submodular
    function and a non-negative modular function are described in 
    \cite{Harshaw2019-ib}.
There have been further advances in the field that provide computationally faster 
    algorithms with slightly worse approximation bounds~\cite{Avdiukhin2019-wc}.
While not an exhaustive list of literature in the domain of submodular maximization,
    the scalability and generalizability of our proposed formulation is made possible by the exemplary contributions of the works mentioned above.
\section{Problem Formulation for Multi-stakeholder Fairness}

We focus on a multi-stakeholder system where an e-commerce 
    platform provides recommendations
    of items from sellers to the buyers browsing on the platform.
When a buyer submits a search query on such a platform, the recommender system 
    suggests relevant items.
Without loss of generalizability, we consider each \emph{seller} on the platform as
    a \emph{stakeholder}, and use the two terms interchangeably henceforth.
In the application discussed in Section~\ref{sec:experiments}, we extend the 
    definition of \emph{stakeholders} to include different sources of property
    listings on Zillow, an online real-estate marketplace.
As described in the previous section, the primary objective of our recommender 
    system is to ensure that the percentage of items belonging to each stakeholder in the items recommended
    to a buyer is proportional to the number of items belonging to the particular stakeholder that are relevant to the buyer.
Henceforth, we refer to this objective as the \emph{coverage objective}.
In addition to the coverage objective, a platform can have secondary objectives
    such as minimizing logistical costs, maximizing utility of recommendations, etc., referred to as \emph{auxiliary objectives}.
We aim to optimize the recommender system such that it achieves an optimal 
    trade-off between both of these objectives.

Next, we formally define our problem setup. Let $\buyers = \{1, \cdots, m\}$ be the
    set of buyers and $\items = \{1, \cdots, n\}$ be the set of items.
It should be noted that only a subset of items $\items_b \in \items$ from the 
    universe of candidate items are relevant to each buyer $b$.
We represent stakeholders as a set of sellers $\sellers = \{1, \cdots, t\}$.
For each item $u \in \items$, let $\sellers(u) \subseteq \sellers$ denote the sellers who
    store item $u$ in their inventory.
Similarly, let $U(s) \subseteq \items$ be the set of items in the inventory of a seller $s \in \sellers$.
The goal of our recommendation system is to suggest $k$ items to the buyer that \emph{fairly}
    cover all the sellers, assuming that $k \leq n$.
When serving recommendations, we use a binary variable $x_{u,s}$ to denote whether an item $u$ 
    is recommended to a buyer covers a seller $s$ i.e., $x_{u,s} = 1$ iff $u \in \items(s)$.
    
\subsection{Stakeholder Coverage:}

To define the stakeholder coverage objective, we first formalize the notion of \emph{fair}
    coverage.
\subsubsection{Fair coverage:}
\label{sec:fair_coverage}
A set of $k$ recommendations denoted by $\recommendations_b$ given to a buyer $b$ is considered
    \emph{fair} to all stakeholders if and only if
\begin{equation}
\label{eq:fair_coverage}
    \forall s \in \sellers: \frac{\sum_{u \in \recommendations_b} x_{u,s}}{k} \geq \frac{|\items_b(s)|}{|\items_b|}.
\end{equation}
Equation~\ref{eq:fair_coverage} ensures that the percentage of each seller's inventory 
    included in the recommended items $\recommendations_b$ is at least as high as the percentage of the 
    seller's inventory relevant to the buyer $b$.
Henceforth, we refer to the ratio $\frac{|\items_b(s)|}{|\items_b|}$ as the \emph{fair coverage threshold}
    of seller $s$ for buyer $b$ and denote it by $\delta_{s,b}$.
It should be noted that S{\"u}rer {\etal} \cite{Surer2018-bh} imposes similar \emph{provider constraint} across all the buyers and sellers
    together:
\begin{equation}
\label{eq:provider_constraint}
\forall s \in \sellers: \sum_{b \in \buyers} \bigg(\frac{\sum_{u \in \recommendations_b}x_{u,s}}{m \times k} \bigg) \geq \frac{|\items(s)|}{n}
\end{equation}
Our definition of fair coverage supersedes such a constraint because ensuring fair coverage for each seller
    in recommendations for every buyer query obviously leads to satisfying the \emph{provider constraint} in Equation~\ref{eq:provider_constraint}
    across all buyers and sellers, but not vice versa.
Furthermore, just satisfying \emph{provider constraints} can lead to biased 
    situations where individual buyers are consistently provided low utility recommendations
    in order to satisfy a global coverage constraint for a seller.
Defining fair coverage on individual buyer makes it possible to augment output of 
    any modern personalized recommendation system with our objective in real-time.
    
\subsubsection{Coverage objective:}
Having defined \emph{fair coverage}, we are now in a position to formalize the coverage objective denoted by $F(.)$.
For a set of recommendations $\recommendations_b$, the value of coverage objective is:
\begin{equation}
    F(\recommendations_b) := \sum_{s \in \sellers} \min\bigg(\frac{\sum_{u \in \recommendations_b}x_{u, s}}{k}, \delta_{s,b}\bigg)
\end{equation}
Let us remind ourselves about the submodular functions.
A set function $f : 2^{V} \rightarrow \mathbb{R}$ is \emph{submodular} if for every
    $A \subseteq B \subseteq V$ and $e \in V \setminus B$ it holds that $f(A \cup \{e\}) - f(A) \geq f(B \cup \{e\} - f(B)$.
\lemma{The coverage objective $F(.)$ is a submodular set function.}
\proof{
For a buyer $b$, consider two sets of items $\recommendations_0, \recommendations_1 \subseteq \items_b$ such that $\recommendations_0 \subset \recommendations_1$. 
Consider an item $u_0 \in \items_b \setminus \recommendations_1$.
Without loss of generality, we assume that $\sellers = \{s_0\}$ i.e., there is a single stakeholder to be covered and $u_0 \in \items(s_0)$. This gives rise to two cases enumerated below.\\
\textbf{Case 1:} The set $\recommendations_0$ provides fair coverage to stakeholder $s_0$.
Hence,
$\frac{\sum_{u \in \recommendations_0}x_{u_0, s_0}}{k} \geq \delta_{s_0,b}$. 
Hence, $F(\recommendations_0) = \delta_{s_0, b}$. 
Given that $\recommendations_0 \subset \recommendations_1$, we also know that $F(\recommendations_1) = \delta_{s_0, b}$.
Moreover, $F(\recommendations_0 \cup \{u_0\}) = \delta_{s_0, b}$ and $F(\recommendations_1 \cup \{u_0\}) = \delta_{s_0, b}$.
Hence, $F(\recommendations_0 \cup \{u_0\}) - F(\recommendations_0)= F(\recommendations_1 \cup \{u_0\}) - F(\recommendations_1) = 0$.\\
\textbf{Case 2:} The set $\recommendations_0$ does not cover the stakeholder $s_0$ fairly.
Hence,
$\frac{\sum_{u \in \recommendations_0}x_{u_0, s_0}}{k} < \delta_{s_0,b}$.
Hence, $F(\recommendations_0 \cup \{u_0\}) - F(\recommendations_0) = 1 / k$.
If the recommendations $\recommendations_1$ fairly covers the stakeholder $s_0$, 
$F(\recommendations_1 \cup \{u_0\}) - F(\recommendations_1) = 0$, else, $F(\recommendations_1 \cup \{u_0\}) - F(\recommendations_1) = 1/k$.\\

\noindent Thus, in both cases, $F(\recommendations_0 \cup \{u_0\}) - F(\recommendations_0) \geq F(\recommendations_1 \cup \{u_0\}) - F(\recommendations_1)$. Hence, proved.
}
\subsection{Auxiliary objectives:}
Let us denote an auxiliary objective by $G(.)$. We differentiate the potential auxiliary objectives into two broad categories viz., maximization objectives and the minimization objectives.
\subsubsection{Maximization Objectives}
\label{sec:max_objectives}
Such an auxiliary objective typically involves maximizing an attribute related to the recommended item
    alongside the primary objective of fair multi-stakeholder coverage.
For example, maximization of the utility of recommendations.
All modern personalized recommender systems typically compute a utility score $r_{u, b}$ denoting the
    relevance of item $u$ to a buyer $b$ using various models such as collaborative filtering or a
    content-based recommendation system.
Thus, total utility of recommended items is:
\begin{equation}
    G(\recommendations_b) := \sum_{u \in \recommendations_b}r_{u, b}
\end{equation}
As the utility scores are non-negative and computed beforehand and generally fixed for each item, 
    we can assume that $G(.)$ is a modular set function (equality in the submodular function definition).
Note that the objectives like maximization of non-negative fixed attributes such as profit margins, etc. can be 
    incorporated similarly.
    
\subsubsection{Minimization Objectives}
\label{sec:min_objectives}
Next, we consider the scenario requiring minimizing an attribute related to the recommended items
    while simultaneously attempting to maximize the fair coverage objective.
For example, one can envision a user-oriented goal that recommends items to a buyer that have minimum cost per unit utility.
In contrast to the utility maximization objective above, we would like to minimize the cost per unit utility of recommendations.
Hence,
\begin{equation}
    G(\recommendations_b) := \sum_{u \in \recommendations_b} c_u/r_{u,b}
\end{equation}
where $c_u$ is the cost of an item.

\subsection{Overall Objective:} 
Combining the stakeholder fair coverage objective with the auxiliary objectives above, we obtain the overall
    objective in the form
\begin{equation}
    \mathcal{F}_\alpha(\recommendations_b) := \alpha F(\recommendations_b) \pm (1 - \alpha) G(\recommendations_b)  
\end{equation}    
where $\alpha \in [0,1]$ allows us to control the trade-off between the two objectives.
Note that we add the two objectives when dealing with a maximization auxiliary objective, and take the difference in case of a minimization auxiliary objective.
A reader may correctly wonder the need for differentiation between the maximization and minimization auxiliary
    objectives, since typically the two are interchangeable with a change of sign.
However, in our case, as described in the subsequent sections, the nature of auxiliary objectives affects the
    properties of the overall combined objective $\mathcal{F}_\alpha(.)$, and subsequently the optimization algorithms.
Hence, we differentiate between the two.
\begin{problem}
\label{eq:og_problem}
Given a set of buyers {\buyers}, a set of items {\items}, a set of stakeholders {\sellers} and platform parameter $\alpha$, recommend a set of items ${\recommendations_b^*}$ to each buyer such that:
\begin{eqnarray}
    \recommendations_b^* &=& \argmax_{\recommendations_b \subseteq \items_b} \mathcal{F}_\alpha(\recommendations_b) \nonumber\\
    \textrm{s.t.,}&& |\recommendations_b| = k.
\end{eqnarray}
\end{problem}
\noindent In the following section, we show how existing contributions from the field of submodular optimization
    can be used to optimize the combined objective with strong approximation guarantees. 
\section{Algorithms}

In this section, we describe the two principal algorithms that we use to optimize the overall objective formulated above.

\subsection{Maximization Auxiliary Objectives}
First, we account for the case where the auxiliary objective is maximization of a non-negative modular set function, described in section~\ref{sec:max_objectives}.
It can be trivially shown that the corresponding overall objective function obtained in this situation is a monotone submodular function.
Such a function can be maximized using a simple greedy approach in Algorithm~\ref{alg:greedy} that builds the recommended items set iteratively.
Let us denote the marginal gain of adding a new item to a set of recommendations by
$\mathcal{F}_\alpha(u | \recommendations_b) = \mathcal{F}_\alpha(\recommendations_b
\cup \{u\}) - \mathcal{F}_\alpha(\recommendations_b)$.
In each iteration, the greedy Algorithm~\ref{alg:greedy} simply adds to the set the item with the largest marginal gain.
Moreover, the solution provided by the greedy algorithm satisfies strong approximation guarantee.
\[
    \mathcal{F}_\alpha(\recommendations_b) \geq (1 - \frac{1}{e})\mathcal{F}_\alpha(\recommendations_b^*)
\]
where $\recommendations_b^*$ is the optimal solution for the Problem~\ref{eq:og_problem} with a maximization auxiliary objective.
\begin{algorithm}
\textbf{Input}: Set of relevant items for buyer $\items_b$, fair coverage
thresholds $\delta_{s,b}$ for each seller, $k, \alpha$\;
\textbf{Output}: Set of recommended items $\recommendations_b$\;
$\recommendations_b = \emptyset$\;
\While{$|\recommendations_b| \leq k $}{
    $z = \argmax_{u \in \items_b}\mathcal{F}_\alpha(u | \recommendations_b)$\;
    $\recommendations_b \leftarrow \recommendations_b \cup \{z\}$\;
    $\items_b \leftarrow \items_b \setminus \{z\}$\;
}
\Return $\recommendations_b$
\caption{Greedy}
\label{alg:greedy}
\end{algorithm}
\noindent Notably, for each buyer $b$, the algorithm iterates through all candidate items on line 5 to find the
    item with the largest marginal gain, resulting in a time complexity of $O(nk)$ per buyer, not accounting for the
    complexity of evaluating the coverage objective itself.
    
In online personalized recommender systems with thousands of potential items, such an approach can potentially
    increase response times, when set of recommendations are updated continuously based on
    user activity within a session.
For such applications, we may achieve a significant speed-up without a loss of quality of the solution by using a 
    priority heap to avoid re-computation
    of marginal gain of all items, as shown in Algorithm~\ref{alg:lazy_greedy}.
    
On account of submodularity, the marginal gain of elements can only decrease in each iteration.
Leveraging this observation, we only need to re-evaluate the marginal gains of a small subset of items per         
    iteration, until the item whose marginal gain computed in the previous iteration is less than the updated marginal gain of the top-most element in the priority heap as shown on line 8 of Algorithm~\ref{alg:lazy_greedy}.
It should be noted that the worst case complexity of Algorithm~\ref{alg:lazy_greedy} is the same as that of 
    Algorithm~\ref{alg:greedy}. 
But in most practical applications, it is significantly faster.
\begin{algorithm}
\textbf{Input}: Set of relevant items for buyer $\items_b$, fair coverage
thresholds $\delta_{s,b}$ for each seller, $k, \alpha$\;
\textbf{Output}: Set of recommended items $\recommendations_b$\;
$\recommendations_b = \emptyset$\;
Create maximum priority heap $H$ and push each key $u$ from $\items_b$ with
value $v_u = \mathcal{F}_\alpha(u | \emptyset)$\;
\While{$|\recommendations_b| \leq k $}{
    Pull top key $i$ from the priority heap $H$\;
    Evaluate new marginal gain $\mathcal{F}_\alpha(i | \recommendations_b)$, $\Phi = \{i\}$\;
    \While{True}{
        Pull top key $j$ from the priority heap with value $v_j$\;
        \If{$\mathcal{F}_\alpha(i|\recommendations_b) \geq v_j$}{
            break\;
            }
        \Else{
            $\Phi \leftarrow \Phi \cup \{j\}$\;
            }
        }
    $z = \argmax_{i \in \Phi}\mathcal{F}_\alpha(i | \recommendations_b)$\;
    $\recommendations_b \leftarrow \recommendations_b \cup \{z\}$\;
    $\Phi \leftarrow \Phi \setminus \{i\}$\;
    \For{each $j \in \Phi$}{
        Re-insert key $j$ into the heap $H$ with value $v_j = \mathcal{F}_\alpha(j
        |\recommendations_b)$\; 
        }
}
\Return $\recommendations_b$
\caption{Lazy Greedy}
\label{alg:lazy_greedy}
\end{algorithm}

\subsection{Minimization Auxiliary Objectives}
Next, we describe the algorithm for the situation where the auxiliary objective is a minimization
    of a non-negative modular set function, described in Section~\ref{sec:min_objectives}.
The main challenge in such cases is that the function $\alpha F(\recommendations_b) -         (1-\alpha)G(\recommendations_b)$ can be either positive or negative, making the overall objective function
    non-monotone.
However, very recent work of Harshaw {\etal}~\cite{Harshaw2019-ib} provides us with theoretically guaranteed fast
    algorithms for such an objective.
The reason a standard greedy algorithm fails to optimize such an objective is described below.
Suppose there is a `bad item' $u$ which has highest overall marginal gain $\mathcal{F}_\alpha(u | \varnothing)$ and so is added to the recommended items set; however, once added, the marginal gain of all remaining items drops below their
corresponding auxiliary objective value, and so the greedy algorithm terminates.
This is sub-optimal when there are other elements $v$ that, although their overall
marginal gain $\mathcal{F}_\alpha(v | \recommendations_b)$ is lower, have much higher ratio between the coverage objective and the auxiliary objective.

To resolve this issue, Harshaw {\etal} use a distorted greedy criterion as shown in line 5 of 
    Algorithm~\ref{alg:distorted_greedy} which gradually places higher relative weight on the stakeholder coverage objective when compared to the auxiliary objective as the algorithm progresses.
However, it should be noted that since the overall objective can be negative, we only recommend items with a     
    positive distorted gain as shown in line 6 of the algorithm.
Hence, for certain values of $\alpha$, we may encounter a situation where the recommendations provided by the 
    algorithm are less than $k$. 
Using Theorem 3 of Harshaw {\etal}~\cite{Harshaw2019-ib}, it can be shown that Algorithm~\ref{alg:distorted_greedy} provides a solution $\recommendations_b$
    for each buyer such that,
\begin{equation}
\mathcal{F}_\alpha(\recommendations_b) \geq (1 - \frac{1}{e})\alpha F(\recommendations_b^*) - (1-\alpha)G(\recommendations_b^*).
\end{equation}
Intuitively, this guarantee states that the value of overall objective is at least as much as would be obtained
    by recommending items of the same cost as the optimal solution, while gaining at least a fraction $(1 - 1/e)$ of its stakeholder coverage.
Furthermore, Algorithm~\ref{alg:distorted_greedy} time complexity can be improved by sampling the items from
    which the best item is chosen in each iteration (line 5).
Theorem 4 of Harshaw {\etal}~\cite{Harshaw2019-ib} shows that, if we sample uniformly and independently
    $\ceil[\Big]{\frac{|\recommendations_b|}{k}\log\big(\frac{1}{\epsilon}\big)}$ items from $\items_b$ in each iteration to maximize over, we can reduce the time complexity to $O(n \log(1/\epsilon))$ where $\epsilon$ is an error 
    parameter, 
    while achieving the same performance guarantee in expectation.
\begin{equation}
\mathbb{E}[\mathcal{F}_\alpha(\recommendations_b)] \geq (1 - \frac{1}{e})\alpha F(\recommendations_b^*) - (1-\alpha)G(\recommendations_b^*).
\end{equation}
The sampling version of Algorithm~\ref{alg:distorted_greedy} is referred to as `Stochastic Distorted Greedy'.
\begin{algorithm}
\textbf{Input}: Set of relevant items for buyer $\items_b$, fair coverage
thresholds $\delta_{s,b}$ for each seller, $k, \alpha$\;
\textbf{Output}: Set of recommended items $\recommendations_b$\;
$\recommendations_b = \emptyset$\;
\For{$i=0$ to $k-1$}{
    $z = \argmax_{u \in \items_b}\bigg\{\big(1 - \frac{1}{k}\big)^{k -
    (i+1)}\alpha F(u| \recommendations_b) - (1 - \alpha)G(u)\bigg\}$\;
    \If{$\bigg\{\big(1 - \frac{1}{k}\big)^{k - (i+1)}\alpha F(u | \recommendations_b)
    - (1 - \alpha)G(u)\bigg\} > 0$}{
    $\recommendations_b \leftarrow \recommendations_b \cup \{z\}$\;
    $\items_b \leftarrow \items_b \setminus \{z\}$\;
    }
}
\Return $\recommendations_b$
\caption{Distorted Greedy (Harshaw {\etal}~\cite{Harshaw2019-ib})}
\label{alg:distorted_greedy}
\end{algorithm}

\subsection{Multiple auxiliary objectives}
One may envisage an application where there are multiple auxiliary objectives.
Algorithm~\ref{alg:lazy_greedy} can optimize multiple maximization objectives together. 
On the other hand, Algorithm~\ref{alg:distorted_greedy} can optimize multiple minimization objectives 
    simultaneously.
Combination of different maximization and minimization objectives together poses an interesting dilemma.
For example, consider an auxiliary objective of the form:
\[
    G(\recommendations_b) = \beta_0 \sum_{u \in \recommendations_b} r_{u,b} + \beta_1 \sum_{u \in \recommendations_b} (-c_u) + \beta_2 \cdots
\]
where $\beta_i \in [0,1]$ controls the relative importance of the auxiliary objectives.
In applications where the individual auxiliary objectives are
    fixed (i.e. known in advance), 
    the operator can verify in ahead of time if $G(\recommendations_b)$ is non-negative and use the appropriate algorithm.

An interesting situation arises if the auxiliary objective value for each item is 
    not fixed.
For the purpose of brevity, we do not discuss such a situation in this work.
However, an enterprising reader may refer to the \emph{Continuous Greedy} algorithm proposed by              
    Feldman~\cite{Feldman2018-vi}.
It maximizes the multi-linear extension of the \emph{coverage objective} alongside an
    arbitrary linear auxiliary objective, and follows it with Pipage rounding procedure to obtain a discrete 
    solution with strong approximation guarantees, as described in Feldman~\cite{Feldman2018-vi}.

% \begin{algorithm}
% \textbf{Input}: Set of relevant items for buyer $\items_b$, fair coverage
% thresholds $\delta_{s,b}$ for each seller, $k, \alpha, \epsilon$\;
% \textbf{Output}: Set of recommended items $\recommendations_b$\;
% $\recommendations_b = \emptyset$\;
% $\varphi \leftarrow \ceil[\Big]{\frac{|\recommendations_b|}{k}\textrm{log}\big(\frac{1}{\epsilon}\big)}$\;
% \For{$i=0$ to $k-1$}{
%     $\hat{\items}_b \leftarrow$ Sample $\varphi$ elements uniformly and independently from $\items_b$\;
%     $z = \argmax_{u \in \hat{\items}_b}\bigg\{\big(1 - \frac{1}{k}\big)^{k -
%     (i+1)}\alpha F(u| \recommendations_b) - (1 - \alpha)G(u)\bigg\}$\;
%     \If{$\bigg\{\big(1 - \frac{1}{k}\big)^{k - (i+1)}\alpha F(u | \recommendations_b)
%     - (1 - \alpha)G(u)\bigg\} > 0$}{
%     $\recommendations_b \leftarrow \recommendations_b \cup \{z\}$\;
%     $\items_b \leftarrow \items_b \setminus \{z\}$\;
%     }
% }
% \Return $\recommendations_b$
% \caption{Stochastic Distorted Greedy algorithm}
% \label{alg:stochastic_distorted_greedy}
% \end{algorithm}
\section{Data and Experiments}
\label{sec:experiments}

In this section, we begin by describing the data obtained from Zillow, an online real estate marketplace,
    and then we evaluate the performance of our proposed approach.
    
\subsection{Data Description}
Real estate buyers visiting the Zillow website are typically shown a paginated list of recommended 
    property listings satisfying the various filter criteria within the searched region.
Zillow uses its proprietary algorithms to personalize recommendations to potential 
    buyers based on their search criteria.
% As the buyers are browsing various listings, they often change the filters on the 
%     viewport.
% As a result, any personalized recommender system tool utilized by the platform 
%     needs to be capable of
%     reflecting these changes in buyer criteria in real time.
The top listings recommended to the potential buyers are typically obtained from various sources.
Majority of the properties on the platform are listed by independent third-party 
    realtors, alongside `New construction homes' listed by
    builders, as well as Zillow owned homes.
Furthermore, some of these listings have various attributes such as availability of     3D or video tours available to the potential buyers.
It is in this context that Zillow is faced with the multi-stakeholder recommender system.

In this application, we consider a random sample of over 13,000 search sessions 
    obtained from buyer interactions on the website.
After applying the filters set by the buyers during these search sessions we end up
    with a collection of over 36,000 potential candidate listings.
In this setting, we consider 5 different stakeholders viz., \emph{independent 
    listings, new constructions, Zillow owned, 3D homes} and \emph{video tours.}
It should be noted that a single listing can potentially belong to multiple stakeholders. 
For example, a new construction home listing may sometimes have an accompanying 3D 
    home tour.
Each listing has a fixed dollar cost and an associated buyer specific utility 
    score.
As a platform operator, our objective is to recommend each buyer 20 listings 
    that cover the relevant
    stakeholders \emph{fairly} as defined in Section~\ref{sec:fair_coverage}.

\subsubsection{Experimental Settings:}
For all the experiments, we run a multiprocessing Python implementation of the algorithms 
    where each process independently
    recommends items for an individual buyer.
All the results presented below reflect the evaluation of the objectives based on a     uniform random sample of 1000 sessions and their associated listings.
Although our application has a limited number of stakeholders, as the computation of 
    the coverage objective is linear in the number of stakeholders,
    our approach does not face scalability issues in situations with a large number of
    stakeholders.
Furthermore, we do not require any extra storage for Algorithm ~\ref{alg:greedy} and Algorithm ~\ref{alg:distorted_greedy}.
In case of Algorithm ~\ref{alg:lazy_greedy}, the extra storage required for the maximum priority heap
    is $O(n)$.
    
\subsection{Experimental Results}
We empirically evaluate the proposed formulation and algorithms for fair multi-stakeholder coverage alongside
    two separate auxiliary objectives viz., utility maximization using Algorithm~\ref{alg:lazy_greedy} and dollar cost per unit utility minimization using Algorithm~\ref{alg:distorted_greedy}.
\subsubsection{Stakeholder coverage:}
In this experiment, we measure the difference between the stakeholder coverage 
    achieved 
    by our approach and the desired target
    coverage required by the \emph{fair coverage} criterion.
\begin{figure}[htp] 
    \centering
    \subfloat[Utility Maximization]{%
        \includegraphics[scale=0.42]{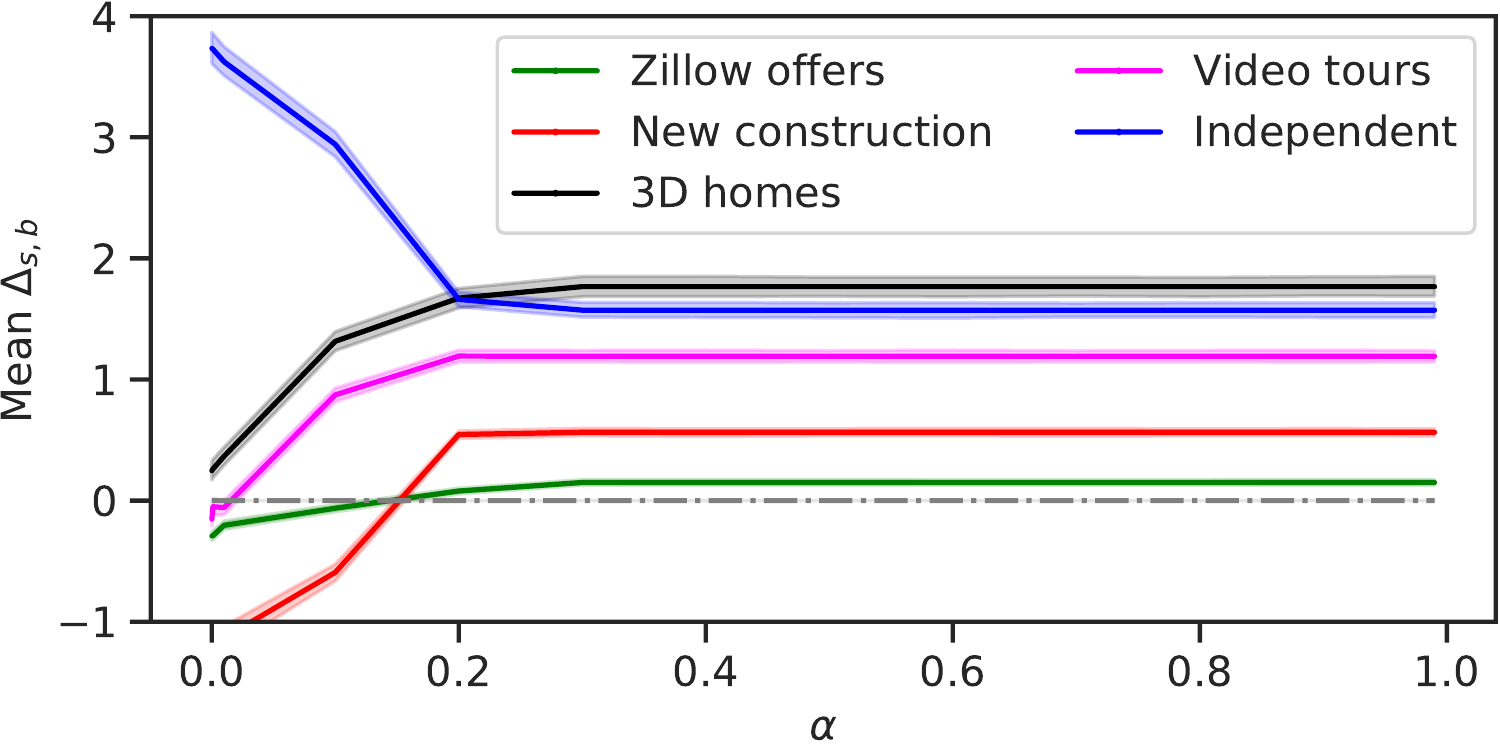}%
        \label{fig:max_score_stakeholder_coverage}%
        }%
    \hfill
    \subfloat[Dollar Cost Per Unit Utility Minimization]{%
        \includegraphics[scale=0.42]{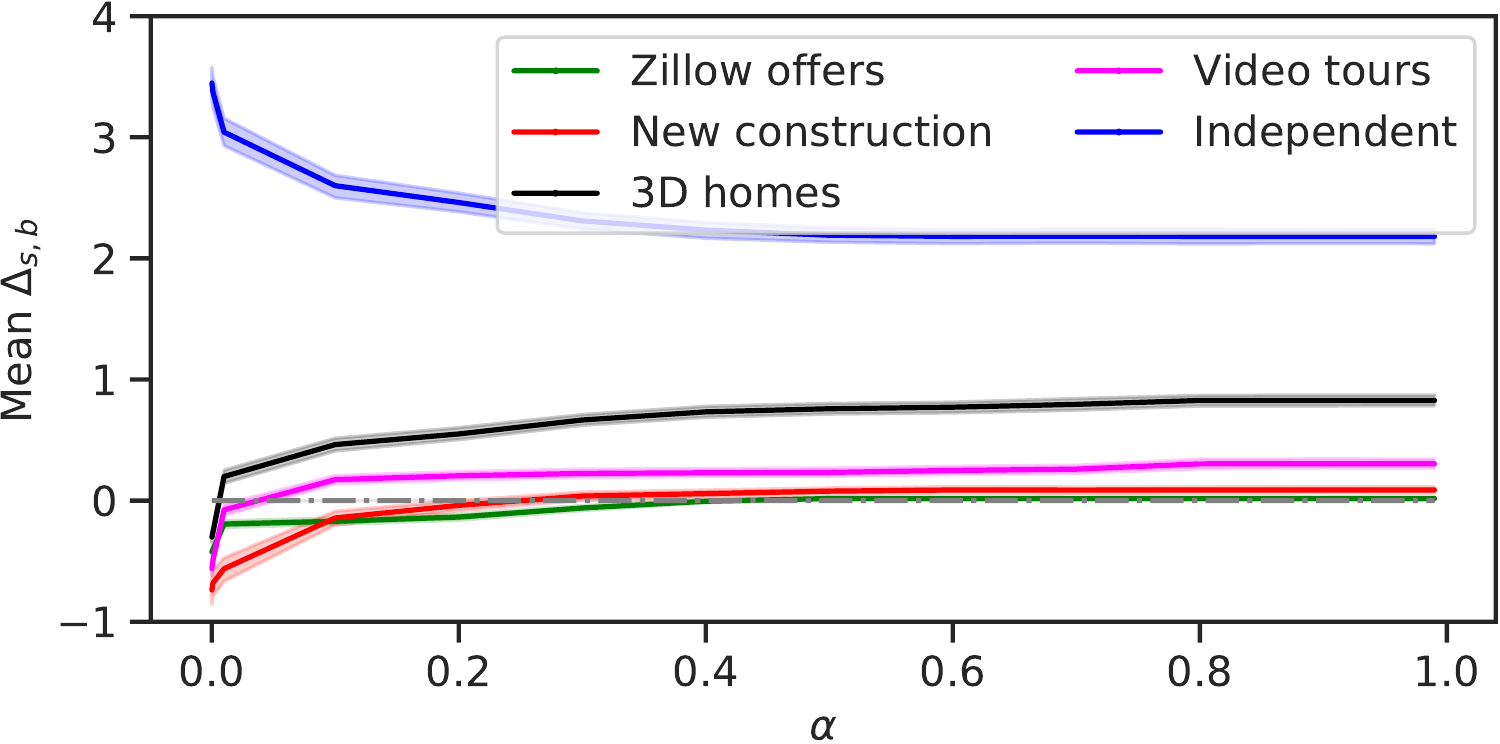}%
        \label{fig:min_cost_stakeholder_coverage}%
        }%
    \caption{Fair coverage of stakeholders}
    \label{fig:stakeholder_coverage}
\end{figure}
\begin{figure}[htp] 
    \centering
    \subfloat[Utility Maximization]{%
        \includegraphics[scale=0.42]{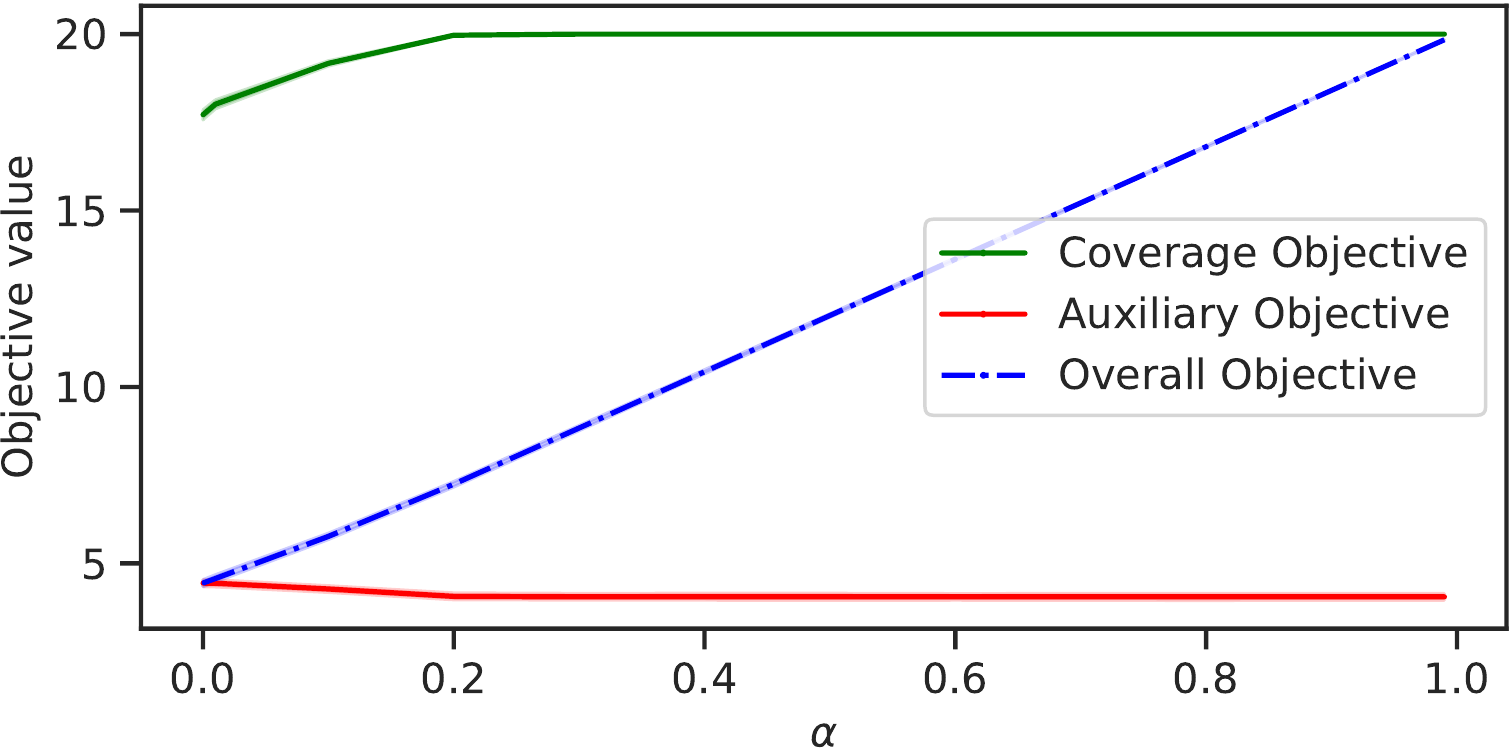}%
        \label{fig:max_score_performance_tradeoff}%
        }%
    \hfill
    \subfloat[Dollar Cost Per Unit Utility Minimization]{%
        \includegraphics[scale=0.42]{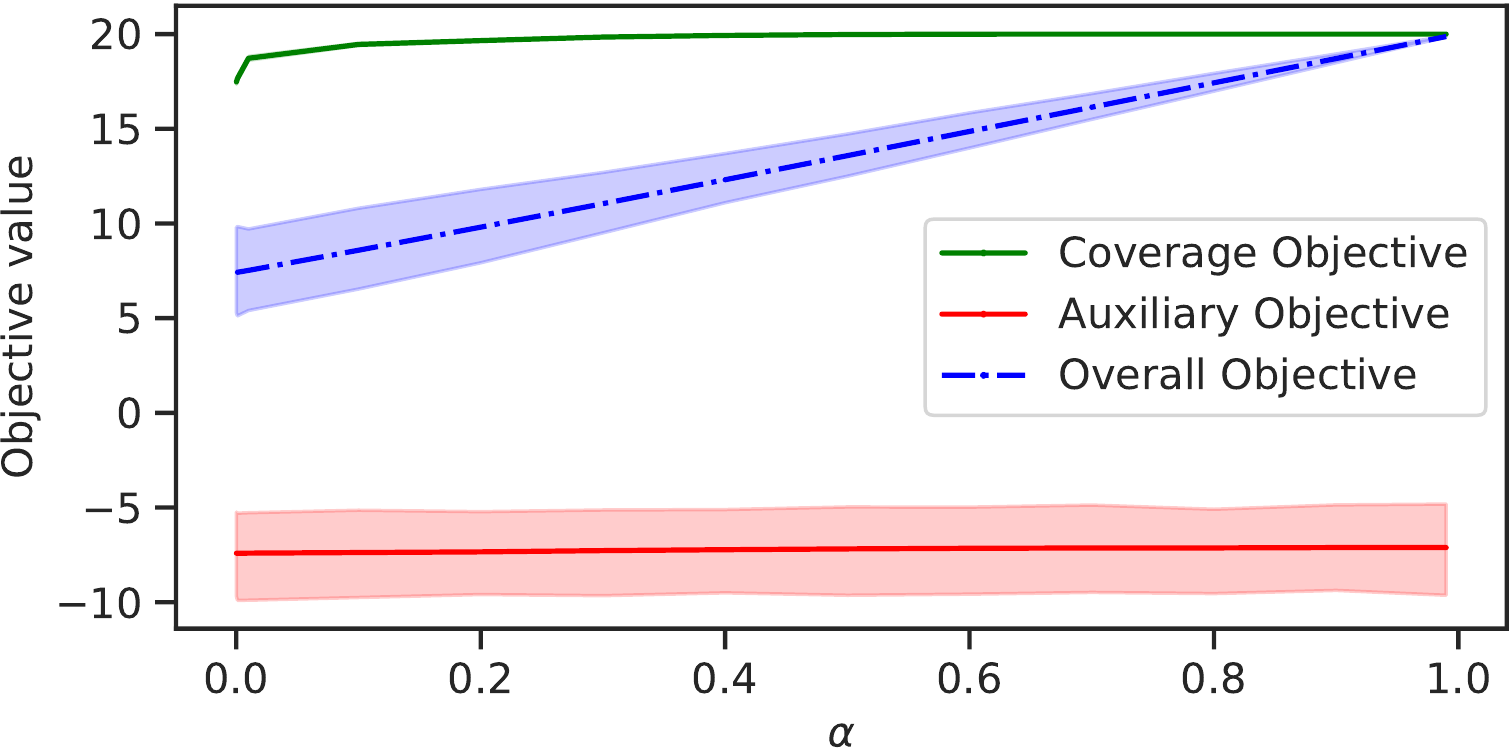}%
        \label{fig:min_cost_performance_tradeoff}%
        }%
    \caption{Trade-offs between coverage and auxiliary objectives}
    \label{fig:performance_tradeoff}
\end{figure}
Specifically, if we represent the proportion of recommended listings belonging to stakeholder $s$ by $\eta_{s,b}$
    and the \emph{fair coverage threshold} for the same stakeholder by $\delta_{s,b}$ as described in Section~\ref{sec:fair_coverage}, then we plot the difference $\Delta_{s,b} = \ceil{k(\eta_{s,b} - \delta_{s,b})}$ averaged over all the buyers, for varying values of the parameter $\alpha$.
When $\Delta_{s,b} \geq 0$ for all stakeholders, we conclude that fair coverage is achieved.
On the other hand, $\Delta_{s,b} < 0$ implies that at least one stakeholder $s$ is under represented in the $k$ 
    recommendations.
In Figure~\ref{fig:stakeholder_coverage}, we observe that a fair coverage of all stakeholder is achieved for both the auxiliary objectives, as the value of $\alpha$ i.e., importance of coverage objective increases.
\subsubsection{Performance trade-offs:}
Here, we visualize the trade-off between the primary fair coverage objective and the auxiliary objective during 
    utility maximization and cost minimization in Figure~\ref{fig:performance_tradeoff}.
For clarity of visualization, we scale the auxiliary objective during cost per unit utility minimization by multiplying it with 
    $10^{-6}$.
We clearly see the trade-off between the coverage objective and the utility of recommendations in 
    Figure~\ref{fig:max_score_performance_tradeoff}.
When $\alpha=0$, only the highest utility listings are recommended.
As the value of $\alpha$ increases, the overall utility of recommended listings is slightly sacrificed in order to improve
    the stakeholder coverage.
During the cost per unit utility minimization in Figure~\ref{fig:min_cost_performance_tradeoff}, the increase in cost of recommendations in order to 
    improve the stakeholder coverage
    is not very apparent in this data due to the use of Stochastic Greedy algorithm with error
    parameter $\epsilon=0.1$.
\subsubsection{Runtime comparisons:}
\begin{figure}
    \centering
    \includegraphics[scale=0.42]{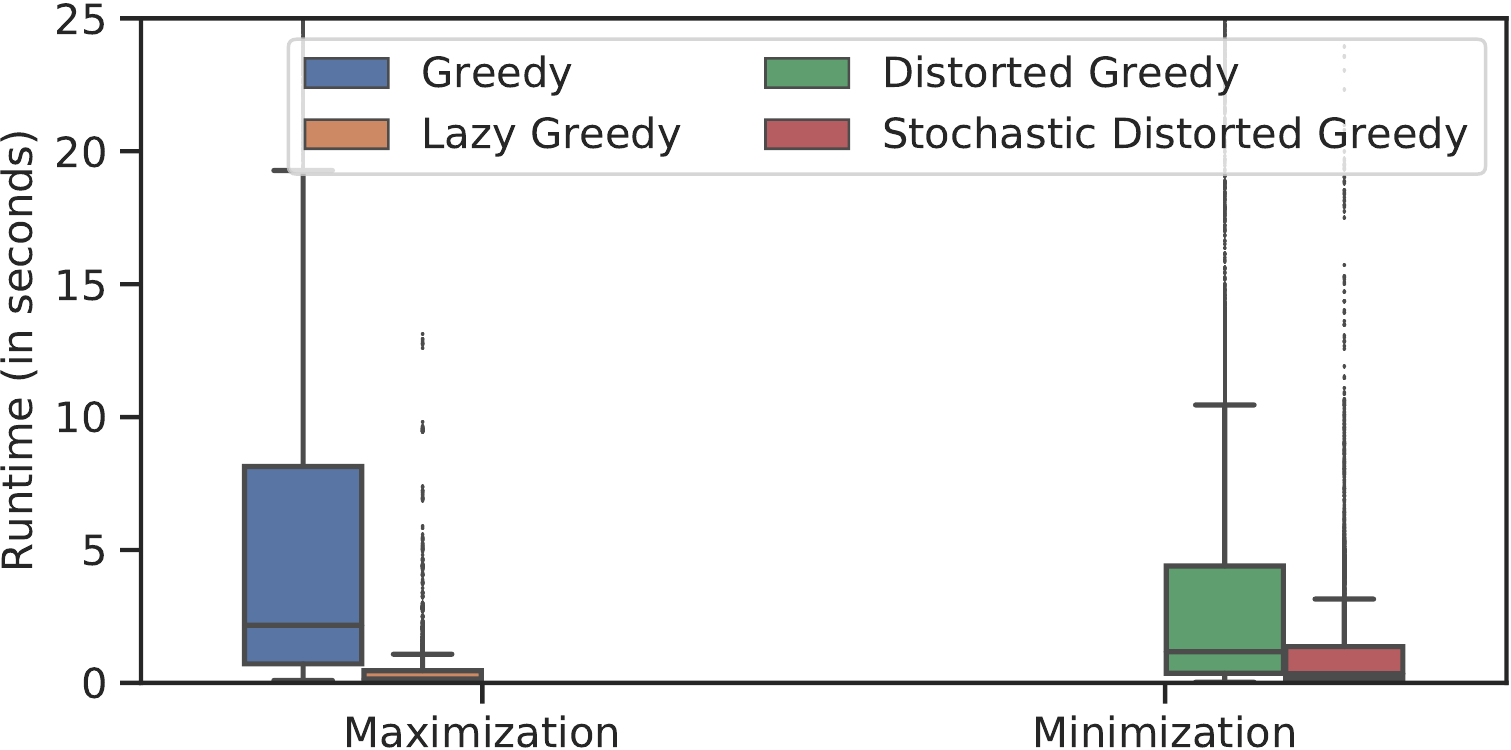}
    \caption{Runtime comparison for auxiliary objectives}
    \label{fig:runtime_comparison}
\end{figure}

Lastly, we compare the algorithm runtimes per buyer for different objectives in Figure~\ref{fig:runtime_comparison}.
While the worst case time complexity for Lazy Greedy algorithm is same as that of
    Greedy algorithm, we observe that it is significantly faster in practice. 
In case of minimization auxiliary objectives, the Stochastic Distorted Greedy algorithm with
    error parameter $\epsilon=0.1$ is more efficient than the deterministic Distorted Greedy algorithm.
Availability of such fast algorithms allows us to use this formulation of fair 
    multi-stakeholder coverage 
    in real-time personalized recommender systems.
\section{Conclusion}

In this work, we study the problem of fair multi-stakeholder recommendations.
Our work confirms the idea that formulating multi-stakeholder coverage objective
    in form of a submodular function allows us to leverage existing submodular
    optimization techniques 
    that can incorporate commonly used secondary objectives in personalized recommender systems.
Using data from an online  real-estate marketplace, we empirically evaluated the
    efficiency and scalability of our proposed approach.
Incorporating non-linear secondary objectives such as learning-to-rank metrics into 
    this framework remains an open research problem.
\bibliographystyle{ACM-Reference-Format}
\bibliography{8_bibliography}

\end{document}